\documentclass{article}

\PassOptionsToPackage{numbers, compress}{natbib}

\usepackage[preprint]{neurips_2026}
\usepackage{wrapfig}
\usepackage{subcaption}
\usepackage{makecell}

\usepackage[utf8]{inputenc} 
\usepackage[T1]{fontenc}    
\usepackage{hyperref}       
\usepackage{url}            
\usepackage{booktabs}       
\usepackage{amsfonts}       
\usepackage{nicefrac}       
\usepackage{microtype}      
\usepackage{xcolor}         

\usepackage{xcolor}
\usepackage{bm}
\usepackage{amssymb}
\usepackage{booktabs}
\usepackage{capt-of}
\usepackage[most]{tcolorbox}
\definecolor{evblueBox}{HTML}{F0F7FF}
\usepackage[table]{xcolor}
\usepackage{multirow}
\usepackage{tikz}
\usepackage{pgfplots}
\pgfplotsset{compat=1.18}

\newcommand{\grayrow}{\rowcolor{gray!20}}
\renewcommand{\arraystretch}{1.5}  

\newcommand{\TODO}[1]{\textcolor{blue}{\textbf{TODO:} #1}}

\title{Neural Events: Discrete Asynchronous Autoencoders for Event-Based Vision}

\author{%
  Roberto Pellerito$^1$ \qquad 
  Daniel Gehrig$^2$ \qquad 
  Shintaro Shiba$^{3,4}$ \qquad 
  Davide Scaramuzza$^1$ \\[2ex]
  $^1$Robotics and Perception Group, University of Zurich \\
  $^2$University of Pennsylvania \qquad
  $^3$The University of Tokyo \qquad
  $^4$Keio University \\[1.5ex]
}
  
\begin{document}

\maketitle

\begin{abstract}




Event cameras capture dynamic scenes with exceptional temporal fidelity by representing them as a continuous stream of microsecond resolution \textit{events}. Each individual event, however, only carries minimal semantic value, merely signaling a localized brightness change. To derive meaningful signals, downstream algorithms need to quickly integrate cues from a potentially massive torrent of low-information events. Current architectures, however, are easily overwhelmed, struggling to balance capturing fine-grained temporal dynamics and maintaining a manageable data throughput. This paper proposes a framework to re-tokenize event streams into a small set of highly informative \textit{neural events}, each representing a local spatio-temporal context window with a discrete learnable code. Every time this code flips, a neural event is triggered, yielding a highly compressed data stream. We demonstrate that, across object detection and classification, networks trained on neural events are on par or surpass the performance of state-of-the-art approaches while reducing the event rate by a factor of 2.0.  
\end{abstract}

\let\thefootnote\relax\footnotetext{
This work was supported by the European Union’s Horizon Europe Research and Innovation Programme under grant agreement No. 101120732 (AUTOASSESS), the European Research Council (ERC) under grant agreement No. 864042 (AGILEFLIGHT), the Swiss AI Initiative by a grant from the Swiss National Supercomputing Centre (CSCS) under project ID a03 on Alps, and by the grant 225354 of the Swiss National Science Foundation.
This work was supported by JST-Research and Development Program for Next-generation Edge AI Semiconductors Grant Number JPMJES2513.
}

\section{Introduction}

Visual information is naturally sparse in time. While much of the world remains predictable, informative signals arise when changes occur, like when a car appears from behind a corner or a water balloon pops. Event cameras capture these salient moments as a stream of per-pixel brightness change measurements, \textit{i.e. } events~\cite{gallego2020event}. Each event captures the pixel location, sign, and timestamp of this brightness change with microsecond latency and temporal resolution, and can thus represent characteristics of highly dynamic scenes (e.g., fast relative motion, highly-textured environments, and/or flickering lights). Individually, however, brightness change events are too fine-grained, carrying little to no meaningful information for downstream tasks; instead, tasks require scalable representations aggregated from patterns of events.

Current approaches to representing event patterns face a fundamental tradeoff: synchronous approaches represent these patterns as dense CNN or Transformer-based feature maps~\cite{zubic2024state, peng2023get, gehrig2023recurrent, li2022asynchronous, perot2020learning} extracted from aggregated and artificially synchronized volumes of events. These feature maps, however, can no longer capture semantic events in an event-driven way. They lack sparsity and precise timing information, leading to wasteful processing, storage, and increased latency. Asynchronous approaches instead capture these patterns as sparse and precisely timed event-based representations, \textit{i.e.} spike trains in Spiking Neural Networks (SNNs)~\cite{cordone2022object, lee2020spike} or spatio-temporal feature graphs in Graph Neural Networks (GNNs)~\cite{schaefer2022aegnn, gehrig2024low}. Event-based representation learning, however, is difficult to scale, especially at high event rates. While GNNs need to construct memory-inefficient event-graphs, SNNs are bound by slow sequential processing and high memory usage. Thus both SNN and GNN-based representations still yield a lower task performance compared to dense feature maps~\cite{hao2025maximizing, hao2026low}.

In this work, we address this gap by introducing a method for representing event patterns as scalable patch-level event representations, \textit{i.e.} \textit{neural events}. Neural events capture precise timing information and are spatially sparse and asynchronous, while summarizing groups of raw and noisy events into high-dimensional feature vectors. At each patch, a Discrete and Asynchronous Encoder recurrently processes event sequences using a linear transformer, producing a sequence of assignment probabilities to discrete codes (\textit{i.e.} tokens) from a learned codebook. Each time the code assignment flips, a neural event is triggered. By using discrete codes as opposed to analog membrane potentials (\textit{i.e.} in SNNs), our method benefits from a low time and memory complexity during training, yielding a low-rate, low-resolution stream of highly informative neural events. To maximize event rate reduction, we propose an unsupervised pre-training strategy which combines reconstruction and temporal smoothness objectives.  Our contributions are summarized as follows:

\begin{itemize}
    \item We introduce an Asynchronous Discrete Encoder for learning \textit{neural events} from raw event streams. Neural events are semantically rich, spatio-temporally sparse, and characterized by their timestamp, patch location, and feature, occupying a compact and discrete latent space.
    \item We introduce a unique tokenization mechanism that triggers a neural event whenever its code flips from its previous state. This strategy can be viewed as a generalization of level-based sampling to the simplex containing all possible code assignment probabilities.
    \item We introduce a pretraining strategy that combines reconstruction and temporal smoothness objectives to stabilize code assignment probabilities. It prevents redundant neural event spiking at decision boundaries, leading to higher event rate reductions and more stable codebooks.
\end{itemize}

We demonstrate that neural events reduce the event rate by a factor of 2.0, while adding only a single bit of extra information per event. 
Despite this reduction, we show across the tasks of object detection (Sec.~\ref{sec:res:detection}) and classification (Sec.~\ref{sec:obj_rec}) that conventional (synchronous) networks trained on neural events perform on par or better than existing synchronous and asynchronous approaches. Since neural events are highly informative, smaller downstream networks can be used, boosting efficiency. Moreover, our encoder only uses 140 KFLOPS per event, $17\times $ lower than the state-of-the-art~\cite{hao2025maximizing}. Finally, we show that, when paired with an asynchronous task network like \cite{gehrig2024low}, neural events can reduce computation while boosting detection performance by 9.0 mAP (Sec.~\ref{sec:obj_det}).

\section{Related Works}

\textbf{Asynchronous Event-based Processing.}
To preserve the microsecond temporal resolution and sparsity of event cameras, several paradigms process event streams asynchronously to avoid dense, frame-like representations.
Early approaches proposed sparse asynchronous convolutions \cite{cannici2019asynchronous, messikommer2020event} to encode the local neighborhood of events, or leveraged PointNet-like architectures to process the stream directly as a point-cloud \cite{sekikawa2019eventnet}. However, these local feature extractors often struggle to efficiently aggregate long-range temporal dynamics, and their computational cost scales poorly during dense event bursts in highly textured, dynamic scenes.

More recently, works like \cite{cannici2020differentiable, santambrogio2024farse} proposed using classical recurrent architectures such as LSTMs to encode raw events, alongside standard attention-based architectures \cite{kamal2023associative, hamaguchi2023hierarchical}.
While effective at capturing temporal context, LSTMs suffer from slow sequential inference over long sequences, and standard attention mechanisms face a quadratic memory and computational bottleneck with respect to sequence length, rendering them prohibitive for high-frequency event streams.
To address sequence scaling, recent frameworks \cite{hao2025maximizing, hao2026low} leverage linear-time recurrent models like RWKV \cite{peng2023rwkv, peng2024eagle}, which admit fast parallel training algorithms based on parallel pre-fix scans, while allowing fast recursive inference at test time.

Other methods treat the event stream as a dynamic spatio-temporal point cloud, forming a graph by connecting events in local neighborhoods and utilizing Graph Neural Networks (GNNs) \cite{gehrig2024low, schaefer2022aegnn, li2021graph, dampfhoffer2025graph}. Despite their high downstream accuracy, GNNs face a critical scalability bottleneck: maintaining and continuously updating complex graph edges for millions of arriving events incurs a massive, unpredictable memory overhead, making them difficult to deploy on memory-constrained hardware.

In contrast to these prior works, our work does not merely map raw events into heavy continuous spaces or graphs. Instead, we leverage linear-time sequence modeling to project events into a \textit{discrete} latent space. This fundamentally solves the bandwidth and memory bottlenecks of prior methods by semantically compressing the stream into compact 7-bit tokens, maintaining the temporal context required for high performance while reducing the downstream computational load.





\textbf{Event Rate Reduction and Encoding.}
As event cameras scale to higher resolutions, the resulting event rates directly inflate storage requirements and saturate real-time transmission bandwidth \cite{gehrig2022highres}.
To counteract this data explosion, various lossless \cite{khan2020lossless} and lossy \cite{banerjee2021lossy} reduction techniques have been introduced.
At the hardware level, modern sensors (e.g., Prophesee Gen4 \cite{finateu20205}) utilize programmable event-rate controllers and complex read-out schemes to throttle data transmission.
These controllers reduce the event rate online by randomly skipping events or dynamically tuning camera biases; however,
they permanently discard high-frequency signals and introduce systematic distortions into the event timestamps.
Denoising filters sample meaningful events and filter out noises,
such as the background activity (BA) filter \cite{Lichtsteiner08ssc,Delbruck08issle} and other spatio-temporal correlation filters \cite{Liu15iscas,Guo20aspdac,Shiba25iccv}.
However, both systematic sampling and denoising remain challenging for practical real-world applications, where it is hard to evaluate these lossy sampling methods.
Notably, since they rely on hand-crafted algorithms and parameters, downstream algorithms must be meticulously co-designed with these heuristic filters to account for the degraded signal, inhibiting generalization.

Instead of hand-crafted destructive filtering, learning-based approaches normally convert the event stream to synchronized representations such as images \cite{Delbruck08issle,lagorce2016hots} or voxels \cite{Zhu19cvpr}.
Further data-rate reduction is proposed, leveraging the sparsity of the voxel representation \cite{sezavar2025learning}.
Our approach is an online, learned semantic association by asynchronously re-tokenizing the event stream to neural events.
Instead of relying on hardware- or software-level filtering, neural events summarize events in a learned way, bypassing distortions while retaining the critical spatio-temporal dynamics required for downstream perception.


\section{Methodology}
\begin{figure*}[t] 
  \centering
  \includegraphics[width=1.0\textwidth]{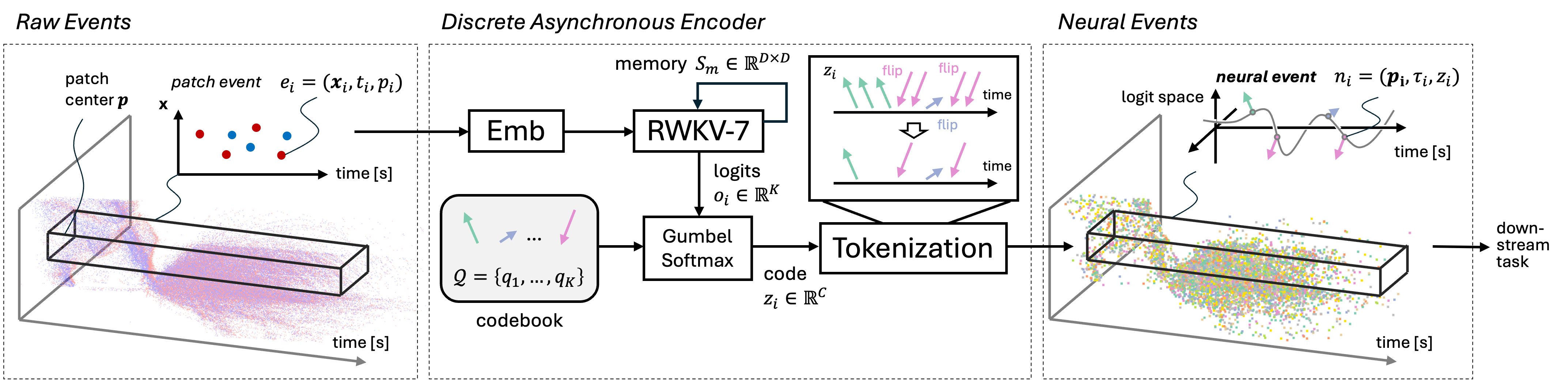}
    \caption{
     \textbf{Method overview.} Our architecture compresses raw events into a low-bandwidth stream of neural events via four stages: ($i$) Raw events $e_i$ are projected into a continuous vector space using spatial and temporal embeddings. ($ii$) 
     A linear-attention sequence model (RWKV-7) updates a localized memory state to output a continuous logit vector $o_i$ per event. ($iii$) A Gumbel Softmax operator maps the continuous logits $o_i$ to a discrete code $z_i$ selected from a pre-learned codebook $\mathcal{Q}$. ($iv$) Unlike encoding, tokenization acts as an asynchronous, many-to-one semantic compressor. It triggers the proposed neural event $n_i$ only when a code flip occurs, i.e., the current event's code differs from the preceding one ($z_{i_m} \neq z_{i_{m-1}}$). If the code does not flip, the redundant event is discarded, preserving semantic context while minimizing data bandwidth. 
    }
  \vspace{-3ex}
  \label{fig:method}
\end{figure*}

We first introduce how neural events are defined, and their analogy to standard events in event cameras (Sec.~\ref{sec:method_overview}) before introducing their generation model (Sec.~\ref{sec:neural_generation}). We then introduce the \textit{Discrete Asynchronous Encoder} at the heart of neural event generation (Sec.~\ref{sec:autoencoder}), and how it is pre-trained (Sec.~\ref{sec:rep_learning}).

\subsection{Method Overview}\label{sec:method_overview}
Event cameras have independent pixels that record ``events'' depending on the scene brightness.
Let an event stream $\epsilon=(e_1,\dots,e_N)$ be composed of a sequence of $N$ events $e_i=(\mathbf{x}_i,t_i,p_i)$. Events are triggered whenever the change in log intensity $\log I(\mathbf{x}_i, t_i)$ at pixel $\mathbf{x}_i$ in  pixel array $\Omega$, and timestamp $t_i$ exceeds the contrast threshold $C$ according to the ideal event generation model~\cite{gallego2020event}:
\begin{equation}
    p_i(\log I(\mathbf{x}_i,t_i)-\log I(\mathbf{x}_i, t_i-\Delta t_i)) = C.
\end{equation}
Here $\Delta t_i$ is the time since the last event at the same pixel and $p_i$ is the polarity (sign) of the brightness change.
Recent event cameras have high spatial resolutions (i.e., $\vert \Omega\vert\sim 10^6$) and thus generate millions of events per second.
In this work, we summarize this high-volume stream of low-information events $\epsilon$, into a low-volume stream of \textit{neural events} $\nu$ via mapping $\phi$:
\begin{equation}
    \phi:\epsilon\mapsto\nu=(n_1,...,n_{N'})\quad \text{where}\quad n_i=(\mathbf{p}_i, \tau_i,z_i) \quad\text{and}\quad \mathbf{p}_i\in \Omega', z_i\in \mathcal{Q}\subset\mathbb{R}^C.
\end{equation}
Each neural event $n$ is characterized by its coordinates $\mathbf{p}$ on a smaller spatial domain $\Omega'$, timestamp $\tau$ and $C$-dimensional feature $z_i\in\mathbb{R}^C$. In particular, we design $\phi$ such that the number of neural events $N'\ll N$, and the domain $\vert\Omega'\vert\ll\vert\Omega\vert$. Furthermore, as shown later, we design $\phi$ such that features $z_i$ only take on a \textit{finite set} of $K$ codes in codebook $\mathcal{Q}=\{q_1,q_2,...,q_K\}$. 

\subsection{Neural Event Generation Model}\label{sec:neural_generation}
Inspired by the working principle of event cameras, we design $\phi$ to act asynchronously, sparsely, and recursively over spatial memory (see Fig.~\ref{fig:method}).
In particular, we tile the image plane into non-overlapping patches with patch centers $\mathbf{p}_j\in \Omega'$ indexed by $j$.
Let $i_1,..,i_{N_j}$ be the event indices in patch $j$ (a total of $N_j$). We implement a recursive module with per-patch memory $S_m\in \mathbb{R}^{D\times D}$ indexed by local time index $m$. Ignoring patch indices for clarity, the update rule is:
\begin{align}\label{eq:overview}
    e_{i_m}', S_{m} = \varphi(e_{i_m}, S_{m-1})\quad \text{with}\quad e_{i_m}'=(\mathbf{p},t_{i_m},z_{i_m}).
\end{align}
The $m^\text{th}$ event in the $j^\text{th}$ patch,  $e_{i_m}$, updates memory $S_{m}$ at patch $j$, and emits modified event $e'_{i_m}$. Note that $e'_{i_m}$ is located at patch center $\mathbf{p}$ and has the same timestamp as the input event $t_{i_m}$ and feature $z_{i_m}\in\mathbb{R}^C$. While the patch-wise design ensures that $\vert \Omega'\vert \ll \vert \Omega\vert$, the resulting events $\epsilon'=(e'_1,...,e'_N)$ still retains the same cardinalty $N$. We thus employ a simple retokenization scheme that only retains events in $\epsilon'$ that significantly differ from their predecessor at the same patch:   
\begin{equation}\label{eq:neural_generation}
    \text{Tokenize}: \epsilon'\mapsto \nu,\quad \text{where} \quad e'_{i_m} \in \nu  \iff \Vert z_{i_m} - z_{\text{ref}} \Vert \geq \theta,
\end{equation}
where $z_\text{ref}$ denotes the feature of the last neural event in $\nu$ triggered at that patch, and $\theta$ denotes the neural contrast threshold.    
As will be shown, restricting the domain of $z_{{i_{m}}}$ to a finite codebook and letting $\theta\rightarrow 0$ yields an efficient $\mathcal{O}(\max_j\log N_j)$ time complexity implementation of Eqs.~\eqref{eq:neural_generation},\eqref{eq:overview} during training, and an $\mathcal{O}(1)$ implementation during testing. This enables more scalable training than classical SNNs, which require $\mathcal{O}(\max_jN_j)$ linear time scans. 

\subsection{Discrete Asynchronous Encoder}\label{sec:autoencoder}
Neural event generation in a single patch follows four steps: \textit{(i)} input projection ($\text{Emb}$), followed by \textit{(ii)} recursive application of linear transformer RWKV-7~\cite{peng2023rwkv}, \textit{(iii)} subsequent quantization ($\text{Quant}$), and finally \textit{(iv)} retokenization. Thus steps (\textit{i}-\textit{iii}) are summarized below:
\begin{equation}
    z_{i_m} = \text{Quant}(o_{i_m})\quad \text{where}\quad o_{i_m}, S^{j}_m = \text{RWKV-7}(\text{Emb}(e_{i_m}), S^{j}_{m-1}).
\end{equation}

\noindent\textbf{Input Embedding.} 
Leveraging the formalism in ~\cite{fang2025event2vec}, $\text{Emb}$ is defined as : 
\begin{equation}
    \text{Emb}(e_i)=\text{Emb}_s(\mathbf{x}_i, p_i) + \text{Emb}_t(\Delta t_i)\quad \text{where}\quad     \text{Emb}_{t}(\Delta t_i)_k =
    \begin{cases}
        \text{sin}\left(\frac{\Delta t_i}{10000^{2k / D}}\right), \text{if k even,}\\
        \text{cos}\left(\frac{\Delta t_i}{10000^{2k / D}}\right), \text{if k odd,}
    \end{cases}
\end{equation}
where we combine the temporal embedding component $\text{Emb}_t$ with the spatial embedding module $\text{Emb}_s$ in \cite{fang2025event2vec} due to its smoothness property with respect to input spatial locations.

\noindent\textbf{Linear Transformer.} In what follows we will set $x_{i_m}=\text{Emb}(e_{i_m})\in\mathbb{R}^D$. Following in the line of linear-time recurrent models \cite{vladymyrov2024linear, gu2023mamba}, we use a 2-block linear-attention operator RWKV-7 \cite{peng2025rwkv}. RWKV-7 mixes temporal features by recursively updating a RWK state $S_m$ at time index $m$, using receptance $r_{i_m} = W_r x_{i_m}$, key $k_{i_m} = W_k x_{i_m}$, value $v_{i_m} = W_v x_{i_m}$ and temporal decay  $w_{i_m}$
\begin{align}
\label{eq:rwkv_state}
    S_m = S_{m-1} \left(\text{diag}(w_{i_m})-\hat{\kappa}_{i_m} (a_{i_m}\odot \hat{\kappa}_{i_m})^\intercal\right) + \tilde{k}_{i_m} v_{i_m}^\intercal.   
\end{align}
Here, $\hat{\kappa}_{i_m},a_{i_m}\in\mathbb{R}^D$ are the removal key, and in-context learning-rate. We use the replacement key $\tilde{k}_{i_m}$ derived from the original key $k_{i_m}$. The output is computed as 
\begin{equation}
    o_{i_m}=W_o(g_{i_m}\odot p_{i_m})\quad\text{with}\quad
    p_{i_m} = \text{LayerNorm}( S_m r_{i_m})+u_{i_m},
    \label{eq:rwkv_output}
\end{equation}
where $u_{i_m} \in \mathbb{R}^D$ is a learned bonus vector that strictly prioritizes the current token, and $g_{i_m}$ is the RWKV gate. Parameters $W_k,W_v,W_r\in\mathbb{R}^{D\times D}$ as well as $W_o\in\mathbb{R}^{K\times D}$ are learnable. 

Processing a new event requires exactly $\mathcal{O}(D^2)$ operations and memory, eliminating the need for quadratic attention and KV-caching in standard Transformers.
During training, the full sequence of events within a temporal patch is known a priori. Memory $S_{m}$ in Equation \ref{eq:rwkv_state} can be efficiently computed over all $k<m$, using a parallel prefix scan \cite{martin2018parallelizing} with $\mathcal{O}(\max_j N_j \cdot D^2)$ complexity, where $N_j$ denotes event sequence length on patch $j$.
As observed by \cite{wang2023state} and \cite{li2022approximation}, memory in linear-recurrent models asymptotically decays exponentially with the number of inputs.
Since our transformer operates patch-wise, this decay is different for each patch, and mitigates the forgetting problem in global event encoding in prior approaches~\cite{hao2026low}.
This, in practice, translates to the model struggling to consistently process events from far-away zones of the event image plane. Our patch-wise design instead constrains attention to the local patches, allowing for longer time modeling and approximate shift invariance.

\noindent\textbf{Quantization.}
Encoding each single event yields continuous-valued vectors $o_i \in \mathbb{R}^{D}$, which increases the information carried by each single event but doesn't provide any temporal, spatial, or memory compression. 
To compress our event stream, we seek to re-tokenize events in a more tractable representation. We thus map $o_i$ to a finite set of codes within codebook $Q\in\mathbb{R}^{C\times K}$, formed out of the elements of $\mathcal{Q}$. As shown in Appendix~\ref{sec:memory_compression}, this allows neural events to only carry 6\% more data than raw events.

We adopt the discrete variational autoencoder formulation ~\cite{RameshPMLR21}, which interprets $o_i$ as logits over the columns of $Q$, and perform straight-through Gumbel softmax assignment  via 
\begin{equation}
    z_i=Qy_i\quad \text{where} \quad y_i=\text{GumbelSoftmax}_\tau(o_i).
\end{equation}
During the forward pass, $y_i$ behaves as a one-hot vector, and during the backward pass, it behaves as a noise-perturbed softmax with temperature $\tau$. This straight-through assignment is analogous to the spiking mechanism in SNNs~\cite{lee2020spike}, handling discontinuities during training.

\begin{figure}[t] 
  \centering
  \includegraphics[width=1\textwidth]{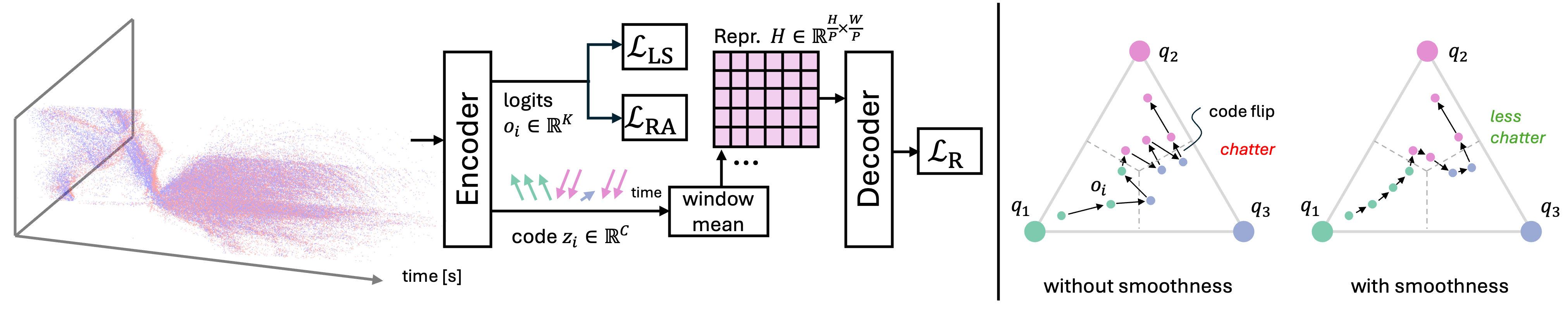}
    \caption{\textbf{Pre-training.} Temporally averaged and stacked codes, $H$ are decoded and supervised with time surface~\cite{sironi2018hats}. To mitigate frequent code flipping (chatter), \textit{i.e.} frequent crossing of the decision boundaries (dashed line) in the probability simplex, a rate alignment and latent straightening loss are applied to the logit sequence $o_i$. This yields smooth trajectories and reduces flipping.}
  \vspace{-2ex}
  \label{fig:straightening}
\end{figure}

\textbf{Event Retokenization.} Here we develop the event trigger condition in Equation~\eqref{eq:neural_generation}. In particular, since the sequence $z_{i_1},z_{i_2},\dots,z_{i_{N_j}}$ from sequence $e'_1,\dots,e'_N$ is piecewise constant, the tokenization scheme can be simplified to 
\begin{equation}\label{eq:simple}
    \text{Tokenize}: \epsilon'\mapsto \nu\quad \text{where}\quad e'_{i_m}\in \nu \iff z_{i_m}\neq z_{i_{m-1}},
\end{equation}
assuming $\theta=0$ and following a quick derivation in Appendix~\ref{sec:app:derivation}. This rule can be interpreted as skipping events whenever there are duplicates, \textit{i.e.} redundant information.  

While retaining equivalence to Eq.~\eqref{eq:neural_generation}, this rule can be implemented with $\mathcal{O}(1)$ time complexity, unlike the standard spiking rule in SNNs, which requires a hysteresis of arbitrary length, and thus needs $\mathcal{O}(\max_j N_j)$ operations.  Moreover, unlike \cite{ohrstrom2025spiking}, which sets a predefined and explicit refractory period, our neural event triggering depends on the dynamics of the scene and is learned.

Our tokenization strategy can be seen as a generalization of level-based sampling to the simplex. Let $\Delta^K:=\{x\in\mathbb{R}^K | \sum_k x_k=1\}$ be the simplex, and $p_i:=\text{Softmax}(o_i)\in\Delta^K$ denote the probability distribution induced by logits $o_i$. Then, let $\delta_k:=\{x\in \Delta^K|\arg\max x = k\}$. Neural events are thus triggered whenever $p_i$ moves between different $\delta_k^K$. We visualize these regions, together with distributions $p_i$ over time in Fig.~\ref{fig:straightening}. Clearly, shaping how these embeddings evolve temporally can have a large impact on the neural event rate. We will discuss next how this representation is learned. 

\subsection{Representation Learning \label{sec:rep_learning}}
We design representations $z_i$ to ($i$) be sufficient to perform useful tasks with downstream networks, and ($ii$) exhibit high compression. We enforce this by pre-training our discrete autoencoder with the following reconstruction and smoothness objectives (see Fig.~\ref{fig:straightening}).
\begin{equation}
    \mathcal{L}=\mathcal{L}_{\text{R}}(\epsilon,\epsilon')+\omega_2\mathcal{L}_{\text{RA}}(\epsilon,\epsilon')+\omega_3\mathcal{L}_{\text{LS}}(\epsilon,\epsilon').
    \label{eq:total_loss}
\end{equation}

\noindent\textbf{Reconstruction Loss.} Our reconstruction loss uses preprocessed events $\epsilon'_t$ in Sec.\ref{sec:autoencoder} and transforms them into a synchronous representation $H$ following the codebook voting strategy, depicted in Fig.\ref{fig:method}. Let $n^k_j$ count the number of events $e'_i$ in patch $j$ associated with code $c_k$. We form a set of patch embeddings $h_j=\frac{1}{N_j}\sum_{k=1}^K n^k_j c_k$ averaging codes according to their frequency in each patch, and concatenate them to an image $H$. We supervise the decoded image with time-surface $\mathcal{T}(\epsilon)$
\begin{equation}
    \mathcal{L}_{\text{R}}(\epsilon,\epsilon')=\omega_0\Vert \mathcal{T}(\epsilon) - D_\phi(H)\Vert_2^2 + \omega_1\Vert\mathcal{T}(\epsilon) - D_\phi(H)\Vert_1.
\end{equation}
While this objective forces neural event representation to be informative, they are not necessarily smooth, and may therefore lead to frequent code flipping (chatter, see Fig.\ref{fig:straightening}). To mitigate this, we employ two temporal smoothing losses, inspired by previous work~\cite{wang2026temporal}.

\textbf{Rate Alignment Loss.} As subsequent events with a similar event rate tend to be emitted by the same source in space, they should possess a similar representation. We leverage this intuition to align the logits $(o_{i_m}, o_{i_{m-1}})$ of subsequent events in the same patch, to their event rate $r_{i_m}$:
\begin{equation}
    \mathcal{L}_{\text{RA}} = \sum_{j} \sum_{m=2}^{N_j}  \Vert \Delta o_{i_m}\Vert^2e^{-\gamma\vert r_{i_m}-r_{i_{m-1}}\vert}\quad \text{with} \quad \Delta o_{i_m}=o_{i_m}-o_{i_{m-1}},
    \label{eq:rate_alignment}
\end{equation}
where $r_{i_m}:=\frac{1}{\Delta t_{i_m}}$ and $\gamma$ is a decay hyperparameter.
This loss forces latent representations corresponding to events with small rate differences to be similar.

\textbf{Latent Straightening Loss.} Code assignment when the codebook distribution is uniform is often noisy, generating numerous code flips at the boundaries of code assignment zones. As depicted in Fig.~\ref{fig:straightening}, the path of the logits $o_i$ often crosses between two code zones. With temporal straightening~\cite{wang2026temporal}, we minimize the velocity angular difference of subsequent logits $o_i$:
\begin{equation}
    \mathcal{L}_{LS} = \sum_{j} \sum_{m=3}^{N_j}  1-\frac{\Delta o_{i_m}^\intercal \Delta o_{i_{m-1}}}{\Vert \Delta o_{i_m}\Vert\Vert \Delta o_{i_{m-1}}\Vert}.
    \label{eq:straightening}
\end{equation}

\section{Experiments \label{sec:experiments}}

We validate our method on the tasks of object detection (Sec.~\ref{sec:obj_det}), and object classification (Sec.~\ref{sec:obj_rec}), showing that our encoder can reduce the original event rate while maintaining highly informative neural events. In Sec.~\ref{sec:res:ablation} we study the impact of codebook size, code dimension, patch size, and pretraining on event rate reduction, computation, and detection performance.

\textbf{Metrics.} Object detection performance is measured in terms of mean Average Precision (mAP), while classification reports top-1 accuracy. Furthermore, we report the MFLOPS/ev and $\mu$J/ev following \cite{gehrig2024low}. MFLOPS/ev count the computation for each new event; whenever a new event arrives, the synchronous method updates the entire representation, while the asynchronous method only needs to perform local updates. $\mu$J/ev instead measures the computation-induced energy, as measured by \cite{jouppi2021ten}. 

\textbf{Training details.} During pretraining, we reconstruct 2-channel time surfaces from $50$-ms event slices from the Gen1 dataset. We set input embedding dimension $D=64$, codebook size $K=64$, code dimension $C=128$, and split the image into patches of $4\times5$ pixels, resulting in $H/4 \times W/5$ patches.
We use a 2-layer RWKV-7 encoder, with eight heads. The weights of the loss in Eq.~\ref{eq:total_loss} are: $\omega_1=1.0, \omega_1=0.1, \omega_2=\omega_3=0.01$. We pre-train for 100 epochs with a learning rate of $10^{-4}$, batch size of 256, and Adam optimizer~\cite{kingma2014adam}. We then freeze our RWKV-7 encoder and codebook (fixing tokenization), while fine-tuning a task decoder (Swin-T) paired with the YOLOX detection heads~\cite{yolox2021} for additional 100 epochs. We train each step using 64 NVIDIA GH200 GPUs. For reconstruction, we reach convergence after 2 hours, while Gen1 requires $12$h. 

\textbf{Datasets.} We report object detection performance on DSEC-Detection~\cite{gehrig2021dsec}, Gen1 \cite{de2020large_gen1}, and N-Caltech101 \cite{orchard2015converting}. DSEC and Gen1 are automotive datasets with ego-centric event cameras capturing a variety of high-speed and stop-and-go motions, while N-Caltech101~\cite{orchard2015converting} comprises short sequences recorded on a pan-tilt device observing image projections on a wall, with associated class labels and bounding boxes. We report classification results on N-Caltech101. For more details, see Appendix~\ref{sec:app:datasets}.

\textbf{Baselines.} For object detection on DSEC-Detection, we compare our approach to the following three synchronous baselines from \cite{gehrig2024low}: ``Inception+SSD'' \cite{iacono2018towards}, ``Events+YOLOv3'' \cite{jiang2019mixed}, and ``Events+YOLOX'' \cite{yolox2021}, which combine images and events. On Gen1 and N-Caltech101, we report a variety of synchronous, asynchronous, and asynchronous to synchronous (A2S) methods like \cite{hao2025maximizing}. A2S methods compute representations event-by-event, while running the backbone synchronously.

\subsection{Object Detection \label{sec:obj_det}}
\label{sec:res:detection}
\begin{wraptable}{r}{0.48\textwidth}
\vspace{-3.0em}
\centering
\scriptsize
\setlength{\tabcolsep}{1pt}
\renewcommand{\arraystretch}{0.95}
\caption{Detection performance on DSEC-Detection (mAP), computation (MFLOPS/ev), and energy ($\mu$J/ev).}
\label{tab:dsec_detection_full}
\resizebox{\linewidth}{!}{%
\begin{tabular}{@{}l cc c c c@{}}
\toprule
\textbf{Model} & \textbf{Images}&\textbf{Async} & \textbf{mAP} & \textbf{MFLOPS/ev} & \textbf{$\mu$J/ev} \\
\midrule
Inception+SSD \cite{iacono2018towards} &\textcolor{green}{$\checkmark$}
& \textcolor{red}{$\times$} & 18.4 & 27,183 & 23,501 \\

\grayrow
Events+YOLOv3 \cite{jiang2019mixed} &\textcolor{green}{$\checkmark$}
& \textcolor{red}{$\times$} & 28.7 & 65,558 & 55,396 \\

Events+YOLOX \cite{yolox2021} &\textcolor{green}{$\checkmark$}
& \textcolor{red}{$\times$} & \textbf{40.2} & 22,049 & 18,631 \\

\midrule
\grayrow
DAGr-S \cite{gehrig2024low} &\textcolor{red}{$\times$}
& \textcolor{green!60!black}{$\checkmark$} & 14.0 & 6.05 & 5.12 \\

\textbf{Ours (TokDAGr-S)} &\textcolor{red}{$\times$}
& \textcolor{green!60!black}{$\checkmark$} & \textbf{25.0} & \textbf{5.81} & \textbf{4.91} \\

\bottomrule
\end{tabular}%
}
\vspace{-1.0em}
\end{wraptable}

\begin{table}[t]
\footnotesize
\setlength{\tabcolsep}{2pt} 
\caption{Detection performance on Gen1 and N-Caltech101 datasets. For MFLOPS/ev $*$ represent lower bound from network backbone and N/A undefined due to spike-based computation. For A2S methods we report the representation complexity and in brackets the end-to-end complexity including the decoder.}
\centering
\resizebox{0.8\linewidth}{!}{%
\begin{tabular}{lccccccc}
\toprule
\multirow{2}{*}{\textbf{Method}} & \multirow{2}{*}{\textbf{Async.}} & \multicolumn{3}{c}{\textbf{Gen1}} & \multicolumn{3}{c}{\textbf{N-Caltech101}} \\
\cmidrule(lr){3-5} \cmidrule(lr){6-8}
& & \textbf{mAP}$\uparrow$ & \textbf{MFLOPS/ev}$\downarrow$ & $\mu$\textbf{J/ev}$\downarrow$ & \textbf{mAP}$\uparrow$ & \textbf{MFLOPS/ev}$\downarrow$ & $\mu$\textbf{J/ev}$\downarrow$ \\
\midrule
Inception+SSD \cite{iacono2018towards} & \textcolor{red}{$\times$} & 30.1 & 27,183 & 23,502 & - & - & - \\
\grayrow
Events+RRC \cite{chen2018pseudo} & \textcolor{red}{$\times$} & 30.7 & >21,758* & >18,386 & - & - & - \\
MatrixLSTM+YOLOv3 \cite{cannici2020differentiable} & \textcolor{red}{$\times$} & 31.0 & >34,519* & >29,168 & - & - & - \\
\grayrow
Events+YOLOv3 \cite{jiang2019mixed} & \textcolor{red}{$\times$} & 31.2 & 65,558 & 55,397 & - & - & - \\
RED \cite{perot2020learning} & \textcolor{red}{$\times$} & 40.0 & 4,712 & 3,982 & - & - & - \\
\grayrow
ASTM-Net \cite{li2022asynchronous} & \textcolor{red}{$\times$} & 46.7 & >21,758* & >18,386 & - & - & - \\

HMNet-L3 \cite{hamaguchi2023hierarchical}& \textcolor{red}{$\times$} & 47.1 & - & - & - & - & - \\

\grayrow
RVT-B \cite{gehrig2023recurrent}& \textcolor{red}{$\times$} & 47.2 & - & - & - & - & - \\

GET \cite{peng2023get}& \textcolor{red}{$\times$} & 47.9 & - & - & - & - & - \\

\grayrow
NVS-S \cite{simonovsky2017dynamic} & \textcolor{green}{$\checkmark$} & 8.60 & 7.80 & 6.59 & 34.6 & 7.80 & 6.59 \\
AsyNet \cite{messikommer2020event} & \textcolor{green}{$\checkmark$} & 14.5 & 205 & 173 & 64.3 & 200 & 169 \\
\grayrow
AEGNN \cite{schaefer2022aegnn} & \textcolor{green}{$\checkmark$} & 16.3 & 5.26 & 4.44 & 59.5 & 7.41 & 6.26 \\
Spiking DenseNet \cite{cordone2022object} & \textcolor{green}{$\checkmark$} & 18.9 & N/A &  - & - &  N/A &  - \\
\grayrow
YOLE \cite{graham20183d} & \textcolor{green}{$\checkmark$} & - & - & - & 39.8 & 3,682 & 3,111 \\


DAGr-S \cite{gehrig2024low}& \textcolor{green}{$\checkmark$} & 30.4 & 4.58 & 3.87 & 70.2 & 6.85 & 5.76 \\

\grayrow
SSLA-L \cite{hao2026low}& \textcolor{green}{$\checkmark$} & 37.5 & 0.72 & 0.61 & - & - & - \\

EVA-L \cite{hao2025maximizing} & \textcolor{black}{A2S} & 47.7 & 2.32 (3,500) & 2.21 (2,958) & - & 2.32 ( - ) & 2.21 ( - )\\
\midrule

\grayrow
\textbf{Ours (TokSwinT)} & \textcolor{black}{A2S} & \textbf{49.9} & \textbf{0.14} (1,130)  & \textbf{0.12} (954) & \textbf{73.0} & \textbf{0.14} (770)  & \textbf{0.12} (210) \\


\bottomrule
\vspace{-3ex}
\end{tabular}}\vspace{-3ex}
\label{tab:flops_det}
\end{table}

\textbf{Setup.} 
We pair our tokenizer with both \textit{synchronous} and \textit{asynchronous} backbones.
The synchronous backbone chains an input projection to 96 dimensions (starting at representation $H$), followed by a Swin-T backbone \cite{liu2022swin} and a YOLOX task head \cite{yolox2021} as introduced by EVA \cite{hao2025maximizing}. 
The asynchronous backbone consists of a state-of-the-art event-by-event processing graph neural network, DAGr \cite{gehrig2024low}. For DAGr, we process the input neural events as a reduced spatio-temporal graph and adapt the input dimension to $C$, the code dimension. We term these TokSwinT and TokDAGr.

\textbf{Results on DSEC.} Tab.\ref{tab:dsec_detection_full} shows that TokDAGr increases the performance of the base model DAGr by $9.0$ mAP, while reducing computation and energy. It thus approaches Events+YOLOv3 without using images, while using roughly $10$k times less computation per event. We argue that the improvement over DAGr comes from its ability to encode temporal histories, which enhance detection, especially in stop-and-go settings.
This result shows that neural events are simultaneously highly expressive and capable of reducing computation through event-rate reduction.

\textbf{Results on Gen1 and N-Caltech101.} Tab.\ref{tab:flops_det} shows the results of TokSwinT. For the A2S models, we separately report the computation and energy required for building the representations and running the entire network (in brackets). 
TokSwinT outperforms synchronous, asynchronous, and A2S models, surpassing the state-of-the-art GET \cite{peng2023get} and EVA-L \cite{hao2025maximizing}. In particular, TokSwinT is more efficient around $17 \times$ (ours 0.14 and EVA 2.32 MFLOPS/ev) at representation building than EVA-L by, as our RWKV-7 uses fewer layers (2 instead of 3) and a smaller hidden dimension ($D=64$) compared to RVT \cite{gehrig2023recurrent} ($D=128$) and EVA-L ($D=192$).

On N-Caltech101, TokSwin-T asynchronous baselines like \cite{gehrig2024low, schaefer2022aegnn} use only 140 $\times 10^3$ KFLOPS/ev to build the representation. The end-to-end computation, in brackets, is significantly higher based on our asynchronous FLOPS measurement protocol, which is required to run a full forward pass for the Swin-T backbone for every tokenized event.

\subsection{Object Classification \label{sec:obj_rec}}
\label{sec:res:recognition}
\begin{wraptable}[10]{r}{0.30\textwidth}
\vspace{-4.0em}
\centering
\scriptsize
\setlength{\tabcolsep}{1pt}
\renewcommand{\arraystretch}{0.95}
\caption{Object classification results on N-Caltech101.}
\label{tab:ncaltech101}
\resizebox{\linewidth}{!}{%
\begin{tabular}{@{}l c c@{}}
\toprule
\textbf{Model} & \textbf{Async.} & \textbf{Acc.} \\
\midrule
HOTS \citep{lagorce2016hots} 
& \textcolor{green}{$\checkmark$} & 21.0\% \\

\grayrow
NVS-S \citep{li2021graph} 
& \textcolor{green}{$\checkmark$} & 67.0\% \\

AEGNN \citep{schaefer2022aegnn} 
& \textcolor{green}{$\checkmark$} & 66.8\% \\

\grayrow
FARSE-CNN \citep{santambrogio2024farse} 
& \textcolor{green}{$\checkmark$} & 68.7\% \\

Asynet \citep{messikommer2020event} 
& \textcolor{green}{$\checkmark$} & 74.5\% \\

\midrule
EST \citep{gehrig2019end} 
& \textcolor{red}{$\times$} & 83.7\% \\

\grayrow
Matrix-LSTM \citep{cannici2020differentiable} 
& \textcolor{red}{$\times$} & 84.3\% \\

Pillar \citep{fan2025eventpillars} 
& \textcolor{red}{$\times$} & 85.3\% \\

\grayrow
E2VID \citep{rebecq2019high} 
&\textcolor{red}{$\times$} & \textbf{86.6\%} \\

EVA-L \citep{hao2025maximizing}
& A2S & \underline{86.3\%} \\

\midrule
\grayrow
\textbf{Ours} 
& A2S & \underline{86.5\%} \\

\bottomrule
\end{tabular}%
}
\vspace{-1.0em}
\end{wraptable}

\textbf{Setup.} After pretraining for reconstruction and detection on the Gen1 dataset, we finetune our model on N-Caltech101, following the test-split of the work \cite{gehrig2019end}, and report classification scores in Tab.\ref{tab:ncaltech101}. We report results for various Asynchronous (A), Synchronous (S), and Asynchronous-to-Synchronous (A2S) baselines.

\textbf{Results.} Tab~\ref{tab:ncaltech101} shows our method performs on par with several asynchronous~ \cite{santambrogio2024farse, messikommer2020event}, synchronous~\cite{rebecq2019high} and A2S~\cite{hao2025maximizing} baselines; notably, it performs similarly to E2VID, which requires reconstructing images from events with additional latency. It also performs similarly to EVA, which requires more MFLOPS/ev to compute the representation (2.32) compared to ours (0.14), as shown in Tab~\ref{tab:flops_det}.


\subsection{Ablations and Sensitivity Studies \label{sec:res:ablation}}
\begin{figure}[t] 
  \centering
  \begin{tabular}{cccc}
  \includegraphics[height=0.2\textwidth]{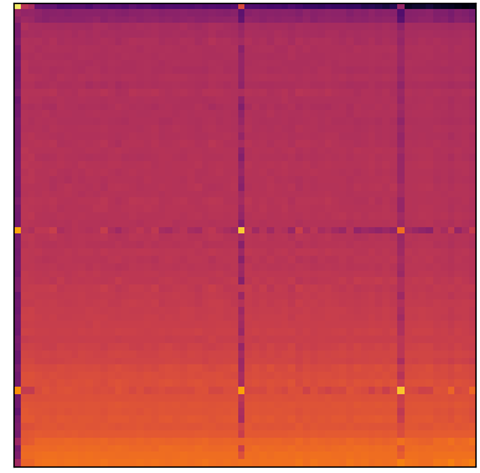}&
  \includegraphics[height=0.2\textwidth]{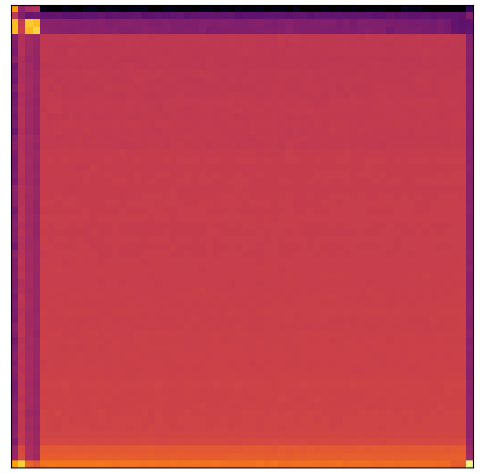}&
  \includegraphics[height=0.2\textwidth]{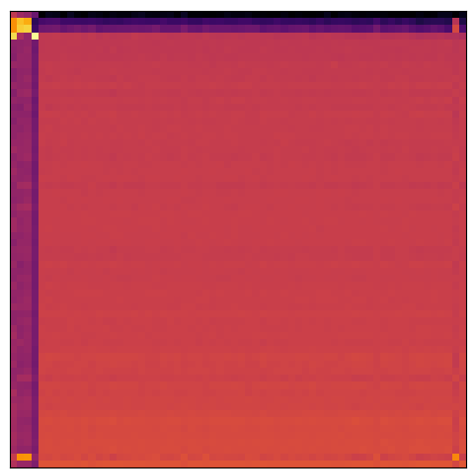}&
  \includegraphics[height=0.2\textwidth]{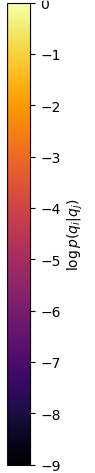}\\
  (a) LS + RA + R&(b) LS + R& (c) R 
  \end{tabular}
  \vspace{-1ex}
    \caption{
   Empirical log probabilities of transitioning from code $i$ to code $j$ for our model trained with  (a) reconstruction (R), rate-alignment (RA), and latent straightening losses (LS), (b) RA + R, and (c) R . Dark lines indicate dead codes. Combining RA and LS losses yields highly non-uniform transition distributions with strong diagonal activations ($i = j$), while omitting them causes codebook collapse (right). Note that low off-diagonal components indicate higher event rate reduction.
    }\vspace{-3ex}
  \label{fig:transition_matrix}
\end{figure}

\textbf{Setup.} We study the impact of ($i$) the loss functions in Equation ~\ref{eq:total_loss}, ($ii$) the patch size, ($iii$) the code dimension $C$ and ($iv$) the code book size $K$. We report the object detection performance on the Gen1 dataset (mAP), the resulting event rate reduction ($R:=N/N'$) in going from raw to neural events, the computational complexity of the encoder, in terms of MFLOPS/ev and the end-to-end latency in ms to process a packet of 100,000 events.
For ($i$), we re-train the different models from scratch on the same split of the Gen1 training set for 40 epochs, and then fine tune the object detector. For ($ii$--$iv$) we train the object detector and representation jointly from scratch. We vary the dimension $C\in \{64,128,256\}$, the number of codes $K\in\{64,128,256\}$ and patch size from $2\times3$, $3\times4$ to $4\times5$. Our base model uses a patch size $4\times 5$, a codebook dimension of $C=128$ and a codebook dimension of $K=64$. For study (\textit{iv}) we set $C=64$.


\textbf{Ablation on loss functions.} Tab \ref{tab:abl:ls_ra} shows that combining both RA and LS losses achieves the highest mAP of 45.8 and a compression rate of $R=2.0$. Removing LS reduces mAP by 12.1 points, while removing RA further reduces mAP by 3.0. Surprisingly, we find that higher event rate reductions indicate higher task performances. We argue that LS and RA are vital to stabilizing the codebook, as shown in Fig.~\ref{fig:transition_matrix}. In each plot we show the empirical log likelihood of transitioning from code $i=1,...,64$ to code $j=1,...,64$.    
While (b-c) show uniform transition probabilities, indicating code collapse, (a) shows a highly non-uniform distribution. Moreover, (a) shows a higher weight on the diagonal, suggesting that RWKV-7 can preserve diversity by selecting many different codes and semantic association by aggregating different events into the same code, avoiding flip where code $i = j$. Finally, combining RA and LS losses reduces dead codes (dark bands).

\begin{table*}[t]
    \centering
    \scriptsize
    \setlength{\tabcolsep}{2pt}
    \renewcommand{\arraystretch}{1.05}
    \captionsetup[subtable]{font=scriptsize, skip=2pt}
\begin{subtable}[t]{0.245\textwidth}
        \centering
        \caption{Smoothness losses.}
        \label{tab:abl:ls_ra}
        \begin{tabular}{@{}cccc@{}}
            \toprule
            \makecell{\textbf{LS}} &
            \makecell{\textbf{RA}} &
            \makecell{\textbf{mAP} $\uparrow$} &
            \makecell{\textbf{R} $\uparrow$} \\
            \midrule
            \textcolor{red}{$\times$} &
            \textcolor{red}{$\times$} &
            30.7 & 1.3 \\
            \grayrow
            \textcolor{green}{$\checkmark$} &
            \textcolor{red}{$\times$} &
            33.7 & 1.6 \\
            \textcolor{green}{$\checkmark$} &
            \textcolor{green}{$\checkmark$} &
            45.8 & 2.0 \\
            \bottomrule
        \end{tabular}
    \end{subtable}
        \begin{subtable}[t]{0.245\textwidth}
        \centering
        \caption{Patch size.}
        \label{tab:abl:patch_size}
        \begin{tabular}{@{}ccc@{}}
            \toprule
            \makecell{\textbf{Patch size}} &
            \makecell{\textbf{mAP} $\uparrow$}&   
            \makecell{\textbf{ms} $\downarrow$} \\
            \midrule
            $2{\times}3$ & 34.8 & 37.9\\
            \grayrow
            $3{\times}4$ & 35.1 & 37.8\\
            \underline{$4{\times}5$} & 43.9  &  36.7 \\
            \bottomrule
        \end{tabular}
    \end{subtable}
    \hfill
    \begin{subtable}[t]{0.245\textwidth}
        \centering
        \caption{Code dimension $C$.}
        \label{tab:code_vectors_dim}
        \begin{tabular}{@{}ccc@{}}
            \toprule
            \makecell{\textbf{$C$}} &
            \makecell{\textbf{mAP} $\uparrow$} &
            \makecell{\textbf{MFLOPS}\textbf{/ev} $\downarrow$} \\
            \midrule
            64 & 40.7 & 0.13 \\
            \grayrow
            \underline{128} & 43.9 & 0.14 \\
            256 & 34.0 & 0.16 \\
            \bottomrule
        \end{tabular}
    \end{subtable}
    \hfill
    \begin{subtable}[t]{0.245\textwidth}
        \centering
        \caption{Codebook size $K$.}
        \label{tab:codebook_size}
        \begin{tabular}{@{}ccc@{}}
            \toprule
            \makecell{\textbf{$K$}} &
            \makecell{\textbf{mAP} $\uparrow$} &
            \makecell{\textbf{MFLOPS}\textbf{/ev} $\downarrow$} \\
            \midrule
            \underline{64} & 40.7 & 0.13 \\
            \grayrow
            128 & 41.8 & 0.43 \\
            256 & 43.3 & 1.38 \\
            \bottomrule
        \end{tabular}
    \end{subtable}
    \hfill

    \caption{Sensitivity of the Discrete Autoencoder on (a) smoothness losses, (b) patch size, (c) code dimension, and (d) codebook size. Underlined is the optimal configuration that we use.}
    \vspace{-4ex}
    \label{tab:ablations_compact}
\end{table*}






\textbf{Sensitivity on patch size}: Tab.~\ref{tab:abl:patch_size} shows that a patch size of $4\times 5$ yields a 9.1 mAP better result than a patch size of $2\times 3$. At the same time, higher patch sizes decrease the end-to-end latency of the autoencoder, as the method needs to process a smaller set of patches.  The latency required to compute the representation for all the patches simultaneously therefore decreases of about 1.2 ms.
Conversely, making the patch larger introduces more variability within the patch, complicating the problem of semantically aggregating events. 

\textbf{Sensitivity on codebook size and code dimension}: 
Tab.\ref{tab:code_vectors_dim} shows that increasing $C$ while fixing $K$ causes a jump in mAP with a modest increase in MFLOPS. Conversely, increasing the $K$ while fixing $C$ in Tab \ref{tab:codebook_size} causes a small increase in performances but a large increase in computation, where doubling the codebook size causes a $\times3$ increase in FLOPS.
Increasing the codebook is beneficial but requires higher dimensional logits, hence more computation. By contrast, using larger code dimensions can lead to code collapse, as more and more noise is encoded into the extra dimensions. Since we are interested in transmitting the codes with as few bits as possible, and low computation, we select $K=64$. Indices within this codebook can be stored and transmitted with 6 bits.

\section{Scope and Limitations}

\label{sec:limitation}
In spite of the computational and bandwidth advantages, our tokenization strategy currently exhibits a performance gap when compared to heavily parameterized, synchronous frame-based methods on complex perception tasks. Furthermore, the reliance on fixed, non-overlapping spatial patches may restrict the model's ability to seamlessly capture highly dynamic objects that cross patch boundaries. To close this accuracy gap, future work will explore dynamic spatial patching and context-adaptive codebooks. Additionally, integrating this asynchronous neural event logic directly at the sensor level, such as within the read-out chip, represents a promising direction for entirely eliminating raw data transmission overhead and realizing natively efficient neuromorphic perception systems.

\section{Conclusion}

We introduce the Discrete Asynchronous Encoder, a novel method for summarizing sequences of raw low-level events into semantically rich neural events. It maps patch event sequences to sequences of assignment probabilities that select discrete codes within a learned codebook. Each time the code flips, a neural event is emitted. Our encoder is $17 \times$ more efficient than state-of-the-art asynchronous event encoders, while producing neural events that are highly informative, and exhibit 2 times lower event rates than the original stream. We show that networks trained with these events are more efficient and perform on par with or better than the state-of-the-art across object classification and detection.  Our work opens the door to more general forms of event representation learning, beyond event-based cameras. We believe that following down this path will be key to enabling ultra-low-power deployment on memory-constrained edge and embedded hardware.


\cleardoublepage
\section{Appendix}
\subsection{Memory Compression \label{sec:memory_compression}}
Typically, raw event representations store the ($\mathbf{x},p,t$) information, with various precision and optimization strategies.
For example, AEDAT~4.0 requires 96 bits (32 bits for ($\mathbf{x},p$) and 64 bits for $t$) \cite{banerjee2021lossy}, and EVT requires $32$--$58$ bits depending on different versions.
Alternatively, a common format that does not store the timestamp is AER (Address Event Representation), encoding each event in 19 bits for ($x,y,p$), however, it is limited in use only for neuromorphic processing  \cite{chakrabartty2023neuromorphic}.

The resulting neural events can be represented by their patch coordinate $\mathbf{p}$, timestamp $t_{i_m}$, and code $z_{i_m}$. Let $b_e$ be the number of bits used to encode a regular event. The total number of bits to encode a stream of $N'$ neural events is then
\begin{equation}
    b_{\text{NE}} \approx N'\left(b_e -\log_2 \frac{\vert \Omega\vert}{\vert\Omega'\vert}+ \log_2 K-1\right)
\end{equation}
where $\log_2 \frac{\vert \Omega\vert}{\vert\Omega'\vert}$ counts the number of bits saved due to having fewer patches than pixels, and $\log_2 K-1$ counts the added bits for the code index in $\mathcal{Q}$, instead of a polarity. This yields a compression ratio
\begin{equation}
    R = \frac{N'}{N}\frac{b_e -\log_2 \frac{\vert \Omega\vert}{\vert\Omega'\vert}+ b_i+\log_2 K-1}{b_e}\approx 1.06\frac{N'}{N} 
\end{equation}
where $N$ counts the number of regular events. We assumed the EVT 3.0 encoding ($b_e=16$ bits),  $\mathcal{Q}$ with a small cardinality  $|\mathcal{Q}| = 64$ and a patch size of $4\times 4$, yielding a $6\%$ bit increase per event. However, as will be shown the event rate reduction $N'/N$ compensates for that, yielding $R<1$.

\subsection{Dataset Details}\label{sec:app:datasets}

\textbf{DSEC-Detection.} The DSEC-Detection dataset \cite{gehrig2021dsec} is a high-resolution automotive benchmark recorded using a Prophesee Gen3.1 event camera, which provides a spatial resolution of $640 \times 480$ pixels. The dataset comprises 53 sequences (roughly 1 hour of total driving time) captured in and around Zurich, Switzerland. It features a wide distribution of challenging lighting and weather conditions, including night-time, day-time, and high-dynamic-range sunset sequences. The dataset provides ground-truth bounding boxes at a high frequency (up to 20 Hz), which are generated by leveraging synchronized RGB images and LiDAR data. It includes annotations for eight distinct traffic participant classes: \textit{pedestrian}, \textit{rider}, \textit{car}, \textit{bus}, \textit{truck}, \textit{bicycle}, \textit{motorcycle}, and \textit{train}, making it one of the most comprehensive benchmarks for urban event-based object detection. 

\textbf{Gen1 Automotive Detection.} The Prophesee Gen1 Automotive dataset \cite{de2020large_gen1} is a large-scale benchmark designed for object detection in realistic driving scenarios. It was recorded using an ATIS-based sensor with a smaller spatial resolution of $304 \times 240$ pixels. Despite the lower resolution, the dataset is massive, containing approximately 39 hours of driving data and over 25.5 million annotated bounding boxes spanning two primary classes: \textit{pedestrians} and \textit{cars}. Gen1 is characterized by a high variance in motion patterns that heavily influence the event generation rate. It features frequent stop-and-go ego-motion (resulting in periods with very few events), a significant component of sensor noise in low-light conditions, and dynamic scenes with traffic participants moving at high relative speeds. This high temporal variance makes it a rigorous testbed for asynchronous and state-based models.

\textbf{N-Caltech101.} Unlike the previous automotive datasets, N-Caltech101 \cite{orchard2015converting} (Neuromorphic-Caltech101) is a standard benchmark for event-based object \textit{classification}. It was generated by capturing the original Caltech-101 image dataset using an ATIS event camera with a resolution of $128 \times 128$ mounted on a motorized pan-tilt unit.   The camera undergoes three distinct, saccade-like triangular motions in front of an LCD monitor displaying the static images. Consequently, the event stream is driven entirely by the artificial motion of the camera over the 2D spatial gradients of the original RGB images. The dataset consists of 8,284 total samples divided into 101 distinct object categories plus a background class. It serves as a fundamental benchmark to evaluate an architecture's ability to extract complex spatial and structural features from event streams generated by consistent, predictable motion.

\subsection{Derivation of Eq.~\eqref{eq:simple}}
\label{sec:app:derivation}

Starting with event generation rule 
\begin{equation}
    \text{Tokenize}: \epsilon'\mapsto \nu,\quad \text{where} \quad e'_{i_m} \in \nu  \iff \Vert z_{i_m} - z_{\text{ref}} \Vert \geq \theta,
\end{equation}
and letting $\theta\rightarrow 0$ yields
\begin{equation}
    \text{Tokenize}: \epsilon'\mapsto \nu,\quad \text{where} \quad e'_{i_m} \in \nu  \iff z_{i_m}\neq z_{\text{ref}}.
\end{equation}
Now let $z_\text{ref}=z_{i_{m-h}}$ be triggered $h$ time steps before $z_{i_m}$. Since $z_{i_m}$ is piecewise constant and no events were triggered between $i_{m}$ and $i_{m-h}$, we can set $z_{i_{m-h}}=z_{i_{m-1}}$, this yields the simplified $\mathcal{O}(1)$ trigger rule in Eq.~\eqref{eq:simple}.

\clearpage
\bibliographystyle{splncs04}
\bibliography{main}


\end{document}